# Modeling and design of heterogeneous hierarchical bioinspired spider web structures using generative deep learning and additive manufacturing

Wei Lu[1,2], Nic A. Lee[1,3], Markus J. Buehler[1,2,4*]

[1] Laboratory for Atomistic and Molecular Mechanics (LAMM), Massachusetts Institute of Technology, 77 Massachusetts Ave., Cambridge, MA 02139, USA

[2] Department of Civil and Environmental Engineering, Massachusetts Institute of Technology, 77 Massachusetts Ave., Cambridge, MA 02139, USA

[3] MIT Media Lab, Massachusetts Institute of Technology, 77 Massachusetts Ave., Cambridge, MA 02139, USA

[4] Center for Computational Science and Engineering, Schwarzman College of Computing, Massachusetts Institute of Technology, 77 Massachusetts Ave., Cambridge, MA 02139, USA

* email: mbuehler@mit.edu

**Abstract:** Spider webs are incredible biological structures, comprising thin but strong silk filament and arranged into complex hierarchical architectures with striking mechanical properties (e.g., lightweight but high strength, achieving diverse mechanical responses). While simple 2D orb webs can easily be mimicked, the modeling and synthesis of 3D-based web structures remain challenging, partly due to the rich set of design features. Here we provide a detailed analysis of the heterogenous graph structures of spider webs, and use deep learning as a way to model and then synthesize artificial, bio-inspired 3D web structures. The generative AI models are conditioned based on key geometric parameters (including average edge length, number of nodes, average node degree, and others). To identify graph construction principles, we use inductive representation sampling of large experimentally determined spider web graphs, to yield a dataset that is used to train three conditional generative models: 1) An analog diffusion model inspired by nonequilibrium thermodynamics, with sparse neighbor representation, 2) a discrete diffusion model with full neighbor representation, and 3) an autoregressive transformer architecture with full neighbor representation. All three models are scalable, produce complex, *de novo* bio-inspired spider web mimics, and successfully construct graphs that meet the design objectives. We further propose algorithm that assembles web samples produced by the generative models into larger-scale structures based on a series of geometric design targets, including helical and parametric shapes, mimicking, and extending natural design principles towards integration with diverging engineering objectives. Several webs are manufactured using 3D printing and tested to assess mechanical properties.

**Significance**: We report a graph-focused deep learning technique to capture the complex design principles of graph architectures – here exemplified for 3D spider webs – and use the model to generate a diverse array of *de novo* bio-inspired structural designs. The work lays the foundation for spider web generation and explores bio-inspired design using rigorous principles. A set of innovative spider web designs is constructed and manufactured, consisting of varied designs with diverging architectural complexity. In future work, this method can be applied to other heterogeneous hierarchical structures including a broad class of architected materials, and hence offers new opportunities for fundamental biological understanding and to meet a set of diverse design opportunities.

## 1. Introduction

Many hierarchical material architectures are found in biological materials[1–5], such as in bone[2,6,7], wood[8], sea sponges and diatoms[9–13], honeycomb structures[14–16], and various spider webs[17–28]. Like all biological structures, spider webs have been well adapted to their natural environments over hundreds of millions of years of evolution[29,30] and a diversity of web structures exist across spider species. The underlying strong silk filament and complex hierarchical architectures ranging from protein scale amino acid chains to macroscale web geometries, provide the structure with (1) exceptional mechanical and biological properties, (2) multiple incorporated functions, and (3) diverse geometries[29,31,32]. With emerging additive manufacturing tools[33–36], it is now possible to generate physical representations of many hierarchical designs; albeit, the synthesis of design solutions remains a challenge, especially when it comes to generating bio-inspired architectures that are rigorously derived from its biological design principles[37–40].

Spider webs (**Figure 1**) exhibit remarkable mechanical properties, including strength, extensibility, and toughness despite their lightweight nature. Moreover, spider silk is biodegradable, biocompatible, and produced from abundant, renewable resources[32]. Spider web structures are also highly adaptive and provide integrated functions such as capturing prey, protection from predators, supporting egg incubation, and structural stabilization under external reactions from wind, projectiles and shifting foundations. *In operando*, spiders modify webs for multiple needs, and self-monitor webs by sensing vibrational information to locate defects and external reactions. These monitoring processes then facilitate prompt responses and proper adjustments to the web architecture[21]. It is known that the primary web architecture and functions are typically constructed within the first few days, and silk is then recycled and produced for structural repairs thereafter[32]. An orb web is the archetypical 2D spider web architecture, which consists of spiral silks and radial threads[41], while 3D webs such as cobwebs, tangle webs, and funnel webs, display more complex structures, higher degrees of randomness, and uncertainty regarding interrelationships among web entities. In addition, there exist local variations with distinct geometries and mechanical properties indicated in 3D spider webs[42]. Both 2D and 3D web geometries naturally vary according to disparities in construction, orientation to the surrounding environment, prey-catching mechanisms[43–45], and defense mechanisms[44–46].

Owing to the exceptional properties and appealing features, spider webs have been researched in various fields, such as in the study of biomaterial platforms[47,48], structural engineering[49,50], battery electrodes[51], and art and music[5,24,26,46]. To increase the accessibility of spider web geometries for various downstream design applications, a digitalization technique has been developed in earlier work to capture and transform the spider web structure to digital graph-structured data[25]. The webs are typically constructed by spiders in a water-trapped cube frame in the laboratory, and then high-resolution 2D images of spider webs are acquired via a sheet laser with a moving-rail set-up and stacked for imaging processing. An image-to-line algorithm translates spider web images into a bead-spring model described in[25] using 3D graph [26]data.

Although spider webs have been imaged in earlier work[25,50], models for the construction of 3D webs do not yet exist, preventing the systematic translation of web design principles into engineering solutions. Since spider webs have a high degree of geometric diversity, and complex variations in scale and functions, it is hard to capture the structure-property relationships through observation-based model building. Although a digitalization technique has been developed for spider web structures[25], high computational costs are required, especially for large-scale 3D webs with more convoluted structures and richer sets of design features.

To address these shortcomings, we propose a deep learning strategy to model complex 3D spider webs, and use the learned representations to construct *de novo*, engineered webs using generative methods. This method innately captures the variations and mutations discovered from synthetic web design that can benefit bio-inspired structural design and design space exploration (**Figure 1a**). While many generative models exist for graphs are often applied to molecular design[52–56], other common issues in current models including graph size constraints, lack of options for training and generating both node labels and edge connectivity, lack of generalizability regarding input data, and limited learning ability for highly complex structures present challenges in the generation of spider web structures. Thus, we propose a new set of deep graph generative models for synthetic web design, with geometric parameters conditioned to better capture the inherent behavior and properties of spider webs, and with permutation invariance considered to improve generation efficiency. Three models are proposed, each with different model architectures (diffusion model, and autoregressive transformer) and different neighbor representations (sparse and full neighbor, or adjacency, matrix), for performance comparison.

Graphs are ubiquitous data representations and suitable for a variety of scientific and engineering tasks, such as molecules, truss structures, social networks, transportation systems, and networks of neurons[57,58]. Graph data is increasingly used for multiple tasks including clustering, classification, regression, and generation in various areas, such as bioinformatics[59–61], natural language processing[62,63], structural design[50,64], and recommendation systems[65,66]. This is mainly due to its potential and scalability to describe complex data structures, the interactions among entities, and information flow for real word systems. Moreover, graphs offer universality and versatility for data implementation among fields[55,67]. Sequences are another flexible data format used in deep learning models, which allows sequence-to-graph data translation but normally requires specific orderings of nodes[68]. Two types of graph data sets are commonly accessed: (1) molecule-centric graph data that consists of atom information (normally type, positions, and charges) and SMILES (Simplified Molecular Input Line Entry System) strings for graph structural representation, such as ZINC, QM9, and GEOM[57,69]; and (2) generic graph data represented by node coordinates and edge connectivity, which we apply here. Other examples also include the *Enzymes*, *Citeseer* datasets, as well as synthetic data sets such as *Grid* and *Ego*[56]. Compared with these data sets, spider web graphs typically have much larger sizes (hundreds to thousands of nodes, tens of thousands of edges), higher structural complexity, and feature a much higher degree of variability of node and edge arrangements.

*Graph Generation Methods*

Synthetic graph generation is a critical task in understanding and modeling complex structures or realistic network systems, discovering new patterns and relationships for structures or materials, and the optimization of structures toward targeted properties[55,56]. During the generation process, graphs are typically learned through the underlying distribution $p(G)$ over graph entities (e.g., nodes and edges), where $p$ and $G$ denote respectively data distribution and graph data input, $G = (V, E)$ specifies the undirected graph, with $V = \{v_1, v_2, ..., v_n\}$ and $E = \{e_{ij} = (v_i, v_j) | v_i, v_j \in V\}$ represent node and edge sets for the graph respectively; $n$ is the total number of nodes, and each edge is connected by node $i$ and $j$. Once models are trained, new graphs are generated using sampling[70]. Traditional graph generation methods that apply mathematical probabilistic models such as Erdős-Rényi, Stochastic Block Models, and Preferential Attachment (PA), are relatively simple to implement but restricted by their learning capacity for realistic graphs[57] that feature intricate relationships and property conditions. Here, deep graph generation models outperform traditional graph generation strategies since they learn the underlying characteristics directly from data and allow larger and more flexible data as input. A general process of deep graph generation typically involves processing data, identifying the model architecture, training, sampling, and lastly evaluating synthetic graphs[56]. We follow a similar strategy here.

Graph generation has wide impacts and applications, and deep learning techniques offer superior learning capacity as mentioned in the previous section. However, challenges exist in training deep models to generate high-quality graphs, owing in part to the size and complexity of graph data, long-range node and edge dependencies, the difficulties of achieving discrete variable learning, the computational demand due to innate node orderings of isomorphic graphs, and limited availability of domain-agnostic data[55,67]. Different types of deep graph generation models have been proposed, each with distinct characteristics and learning capabilities. The types of model architectures that are commonly used are variational autoencoders (VAEs), generative adversarial networks (GANs), autoregressive models (ARs), and diffusion models[56,71]. An illustration for each architecture is shown in **Figure S1**, and existing generators are listed in **Table S1** in the supplementary material. **Section S1, Overview of graph generation models**, in the Supplementary Material section, provides an in-depth discussion of each of the methods.

## 2. Results and Discussion

We first focus our attention on the types of physical graph structures studied here, specifically spider webs. **Figure 1a** shows a summary of the overall approach of digitizing spider web structures and representing them as graphs. Based on data gathered from tomography scans of spider webs[24–26,29,32], we develop a graph model, which is then modeled, and transformed into, new bio-inspired, synthetic structures. As shown in **Figure 1b-c**, the structures feature a high degree of spatial heterogeneity. We show this via coloring, featuring a variation of fiber edge length, clustering coefficient[72], and number of neighbors at each node. **Figure S2** provides additional statistical analysis of key geometric graph properties of the spider webs, complementing the visual analysis shown in **Figure 1b-c**. **Figure S2a** shows the length of each edge, projected to its connected nodes and averaged, and **Figure S2b** shows the clustering coefficient. **Figure S2c** depicts the histogram of the reciprocal density of neighbors (calculated as the distance to each neighbor in the graph divided by the number of neighbors; this means that low values indicate high density). **Figure S2d** depicts the geodesic distance for each node from an arbitrary point on the web's edge[73], which may also provide a measure of signal transduction. Its variance gives a metric connectivity and efficiency of travel across the web (high variance means more localized regions; low variance means a consistent level of density and connectivity). **Figure S2e** shows the normalized second eigenvector statistics of the graph Laplacian projected on a per node basis[74], where more similar values indicate "neighborhoods" or anatomy and high variance indicates more isolated regions of anatomy. These detailed analyses of the structural anatomy of the webs provide key evidence that spider webs are highly heterogeneous, and that in order to understand their construction principles, we must find ways to learn these in appropriate representations.

**Figure 1d-g** shows the development of web construction representations using inductive sampling[75]. Samples of large webs (**Figure 1d**) are developed into a large training set (**Figure 1f**) via inductive sampling (**Figure 1e**) considering up to fourth nearest neighbors, and their length distribution is indicated in **Figure 1g**. **Figure 1h** shows the hierarchical structuring used here, by breaking down a complex graph (Level 4) into a local graph assembly (Level 3), a local web segment as defined by inductive sampling (Level 2), individual silk segments (Level 1, defined by a finite length fiber with a start and end point), and the elementary fiber (Level 0). In the strategy pursued here we seek ways to mimic the natural spider web construction principles up to Level 2 in the hierarchy, but use synthetic strategies to then assemble these structures into bio-inspired materials at Level 3 and 4. It is noted that we also conducted experiments to confirm that the algorithm proposed here can model and generate graph structures of the entire web (as shown, for instance in **Figure 1d**); however, this is of less interest since we want to discover construction principles and then use these to develop synthetic de novo web structures. Moreover, we only have a limited number of unique large webs; hence, not enough data to learn from. This is why we use inductive sampling to capture the spatial heterogeneity identified in **Figure 1b-c.**

Three deep learning models are used in this paper (details see **Materials and Methods**), based on either a diffusion architecture or a generative pre-trained transformer approach. Both models utilize self- and cross-attention as a mechanism to represent long-range feature dependencies. **Figure 2a-d** shows a summary of the diffusion algorithm, which translates noise to graphs (**Figure 2a**). **Figure 2b** shows the U-net architecture used here, along with details on the combined cross-attention architecture to realize conditional denoising (**Figure 2c**) with respect to both, the conditioning variables c and the time step *t*. **Figure 2d** shows the forward Markov and trained reverse Markov chain that is learned by the U-net architecture. **Figure 2e-g** shows a summary of the autoregressive transformer architecture.

Both general modeling strategies use a similar graph representation structure, as summarized in **Figure S3**. Each inductively generated sample is encoded as described in the schematic, using either a full adjacency matrix or a sparse neighbor representation. **Figure S4** provides a summary of the properties used to condition the generative models, featuring a total of 7 geometric properties defined in eqs. (1-4). The plots show histograms of each of the variables. Details on the dataset, including construction and normalization of the data, are included in **Materials and Methods**.

**Figure 3a-c** shows samples of generated web sections using the sparse analog diffusion model (**Figure 3a**), the discrete diffusion model (**Figure 3b**), and the autoregressive transformer architecture (**Figure 3c**). **Figure 3d-e** shows a comparison of the performance of these models respectively, via self-correlation of conditioning variables $C$ (*x*-axis, labeled GT, over the entire test set $\{C_i\}_{\text{test,GT}}$) with measured properties of the generated graphs (*y*-axis, labeled Predicted, $\{C_i\}_{\text{predicted}}$). The resulting R2 values, measuring how the conditioned graph generation parameters relate with measured ones from generated graphs are identified as 0.89, 0.84 and 0.83. The comparison shows that the sparse analog diffusion model has the best performance. It is emphasized that since each of the test samples is unique (since it was sampled from a unique node in the original experimental spider web data), the testing conducted and shown in **Figures 3e-f** are significant in terms of validating the models' predictive generalization capacity. This is because the models have never seen the particular combinations of conditioning

parameters used in these tests, but as the results show, the generative algorithm can successfully produce webs that meet these design demands well.

With these models, we can generate a variety of complex graph sections and use these to assemble larger-scale webs. This are intended to amalgamate natural design principles captured through deep learning with synthetic design objectives; in out study these are mathematically parameterized to yield multi-level architectures. We use generated graph sections to design mathematically parameterized distributions of webs in 3D space, as shown in **Figure 4**. The figure summarizes various aspects of the algorithm and the results of larger-scale generated spider web-based graphs. At the onset, each generative model produces small, local inductive graph sections that have a potentiality to be assembled into larger-scale architectures. There are various dimensions to the process, including how individual sections are integrated. **Figure 4a** shows a shuffling algorithm by which the distance matrix is shuffled symmetrically to obtain more diverse stacking results, as multiple graphs are integrated via operations in the *z*-space of coordinates and adjacency matrix. Indeed, this strategy helps us to randomize the inductively sampled graphs and achieve greater diversity of how nodes of stacked graphs are connected. This becomes clear as we look at **Figure 4b**, which depicts an example of how two identical graphs are stacked, forming a larger graph. In this process, coordinates in the overlapping region are either averaged or taken from the second graph; and coordinates of identical graphs are shifted by *dx, dy* and *dz* (in the example shown on the right, extreme choices of displacements are used to visually represent the new connections, and new graph, formed). **Figure 4c** shows an example of a larger-scale stacking, repeated multiple times, and defining *dx, dy* and *dz* to form a helix. The resulting graph is shown on the right. **Figure 4d** shows another example, here without shuffling, forming a larger helical graph. The center graphs in panels **c** and **d** show the coordinates of the nodes, over node numbers.

Generation of larger synthetic web geometries can be achieved either by using a single graph as a basis or using a set of graphs. Alternatively, we can use the generative models to continuously produce new graphs (either randomly conditioned or conditioned to meet certain target needs). **Figure 4e-f** shows examples of continuously sampled graphs, producing complex architectures. In the two examples, we use the autoregressive transformer model (Model 3, **Figure 4e**) and the analog diffusion model (Model 1, **Figure 4f**) to realize graphs. At each stacking step, a new graph is sampled, conditioned based on a randomly generated conditioning vector *c*. **Figure S5** depicts a gallery of various designs, showing the architectural complexity and variety that can be achieved using the algorithm. Different placement methods are used to shift and transform the positioning of graphs in space at each generative step.

**Figure 5a-b** demonstrates further how spiderweb geometries can be generated along input paths in order to create more complex geometries. For example, parametric equations[76] can be used to create a closed bounding path which is then infilled with web geometries generated using the non-sparse diffusion model. The resulting structure comprises a truss-like path oriented along the bounding curve with connections formed based on the input spiderweb graph information. A closed parametric bounding curve is defined by a set of three equations designating the position of points in cartesian space (details see **Materials and Methods**). These points were then joined into a curve along which a web could be generated. The parametric curve described by this function is shown in **Figure 5a** and loops in on itself in three distinct regions before closing. The output complex web geometry developed by the diffusion model along this path is shown in **Figure 5b**. The generated web formations were meshed and constructed via additive manufacturing in order to observe their behavior as physical objects. **Figure 5c-d** displays two distinct web geometries generated along the same parametric bounding curve. Both objects were manufactured via polyjet 3D printing with a maximum bounding dimension of 15 cm and a strut radius of 0.4 mm (details see **Materials and Methods**).

Helical geometries were additionally generated in order to provide a form that would be suitable for tensile testing. Three distinct helices were used to generate three web patterns which were merged with block regions that could be clamped during mechanical testing. **Figure 6** shows the generated models and manufactured 3D prints during the mechanical testing process (recorded testing videos are provided in **Supplementary Information**, see **Movies M1-M3**). The geometries yielded distinct force-extension behaviors and the impact of individual struts breaking can be observed in the output data. Samples H1, H2 and H3 vary in coil number (2:2.5:4), and sample H3 is designed with a wider-radius web compared with H1 and H2. The strut breakings are observed near the middle along each helix design (**Figure 6b**). Based on the force-strain plot (**Figure 6c**), H1 and H2 show similar stretching behavior, and minor local breaks are noted during the process, which is possibly because the design sections have random fiber segment arrangements, and materials are 3D-printed layer by layer during additive manufacturing. The samples tend to deform until 10% strain (H1 and H2) and up to 25% for H3, indicating that increased deformability can be engineered by designing the helical microstructure. Considering similar cross-section areas of the three samples as indicated in **Figure 6a**, H1 and H2 present higher strength and stiffness than H3, since the underlying web architectures are more densely connected. **Movie M4** shows 3D renderings of the structure shown in **Figure 5a**, as well as H2 and H3; STL files of all five designs are included as **Supplementary Information**). More detailed comparison and analysis could be further anticipated with more systematic tests.

### 3. Conclusion and Outlook

We showed that spider webs are highly heterogenous structures (**Figures 1b-c** and **S2**) that offer complex structural design cues. To capture these, we successfully developed, trained and applied a set of deep generative modeling methods for hierarchical bio-inspired *de novo* spider web designs, based on experimentally scanned and digitalized 3D web structures. Three deep learning models are constructed, analyzed, and compared for spider web generation performance: A sparse analog diffusion model, a non-sparse discrete diffusion model, and an autoregressive generative pre-trained transformer model that also captures the full adjacency matrix. Among these, the sparse analog diffusion model outperforms others in terms of the correlation of graph conditioning variables (**Figures 3d-f**). In other applications, for instance for very large graphs, this model will have additional benefits as it is computationally more effective and requires less memory due to the underlying sparse design. Sparsity is seen in many physical graph structures beyond spider webs, so we foresee many fruitful applications of this method.

We further developed a mathematical parametrized assembling method for bio-inspired graph design, by defining operations in the *z* space of graph representations (**Figure 4**). Based on the generated web sections, we provide a gallery of larger-scale 3D web designs (**Figures 4e-f, S5, 5 and 6**), depicting not only realistic spider web properties but also architectural diversity and complexity (e.g., helix or attractor-like design principles). A selection of these designs is 3D printed with simple mechanical tests conducted, as reported in **Figure 6** (see **Supplementary Movies M1-M4**).

The principal contributions of this work are:

- We generated a spider web data set based on experimental data[25,42,50], processed through an inductive method to provide a diverse collection of graph data that can be used to train the models, using inductive graph sampling[75] (**Figure 1e**). The resulting spider web data has uniform graph sizes, and a set of geometric properties is prepared for conditioning and evaluation (**Figure S4**) to model, capture and synthesize new samples that capture the rich heterogenous feature set that natural webs have.
- The results shown in **Figures 1b-c** and **S2**, and in **Figure S4** for the inductively sampled webs, are important as they provide fundamental insight into the design principles exploited by spiders; revealing that they modulate local structures to achieve a variety of distinct geometric, and by extension, mechanical and other properties in the web. Spider webs are not homogenous structures but consist of a complex assembly of local variations. These natural structural designs are used here to construct bio-inspired structures and could themselves be a subject of further investigation to deeper understand spider web architectures.
- Using the general framework of attention models, we reported three deep graph generative models that show high generalizability and allow flexible data input (**Figure 2**). The models overcome some of the earlier domain-specific limitations and are applicable to other generic graph input data, including for very large graphs due to the efficient computational strategy implemented. The foundation of the models based on attention/transformer strategies intrinsically allows sequences and multiple data types as input, which broadens the data type range and modalities, and can easily be integrated with other language models[77,78].
- We show the general applicability of both, attention-based diffusion models (**Figure 2a-d**) and autoregressive transformer architectures (**Figure 2e-g**), used here to generate diverse and novel web samples. The use of language models offers a generalizable approach towards complex hierarchical system modeling, ranging from language[77] to images[79,80] to dynamical processes[81,82].

- The models show high learning capacity, generalizability and scalability (**Figure 3**). Both node label and edge connectivity of graph data can be learned effectively through training, with conditioning available to integrate graph properties. At the same time, both categorical features (discrete values such as node index and neighbor list) and continuous features (e.g., nodal coordinates) are considered. Due to the computational efficiency, graph generation is possible even for large-sized graph data (e.g., for spider webs: thousands of nodes, tens of thousands of for spider webs).

- The strategies reported here can be applied for data augmentation for spider webs and other large-scale graph data, with lower computational costs than current methods. Moreover, future studies on spider webs could be processed with additional synthetic data combining 3D printing techniques and laboratory mechanical tests, to better understand the structural-property relationships, optimize structural performance, and learn natural design intelligence.

This work lays the foundation for spider web generation and explores bio-inspired design (**Figures 4-6**). A set of innovative spider web designs is constructed, consisting of varied *de novo* designs with diverging architectural complexity. Extending the scope, this method can be applied for other similar hierarchical structures and offers new design opportunities. The approach described here realized a systematic design strategy to capture a series of hierarchical levels (**Figure 1h**) and then generate novel architectural designs. The integrated convolutional-attention approach used in the U-Net architecture of the diffusion model provides the most effective mechanism of capturing multi-level features, and could provide a platform for other similar scientific machine learning applications.

The size of the web samples used to train the model, mimicking features up to Level 2 in the hierarchical makeup, can be adapted by changing the depth of the inductive sampling. By including higher orders of inductive sampling depths, we can generate larger local web samples. Ultimately, these variations can be explored as design principles in the process used here. Other systematic variations of the design process could include the introduction of gradients of properties along a spatial direction, e.g. the helix length axis, to achieve additional material parameter structuring in line with the Universality-Diversity-Principle that provides a platform to achieve distinct mechanical and other properties from the same building blocks; here, elementary silk fibers architected into a variety of forms using the principles outlined in **Figure 1h**. Several explorations of de novo design concepts are shown in **Figures 4-6**, including additively manufactured samples that were exposed to mechanical testing.

Whereas the work report reported here focused on targeting conditioning parameters *C*, all three models allow for flexibility when it comes to generating increasingly diverse or more "creative" samples. For instance, we can use Ancestral Gumbel-Top-k sampling[83], for instance, for the transformer architecture to achieve a greater diversity of sampled webs. In the diffusion model, we can adapt the denoising time-steps and use classifier free guidance[84] during sampling. These strategies have been shown to be effective in other generative models. Another natural dimension that could be explored is masking. All models allow for masking certain regions of an existing web that should or should not be adapted, providing a mechanism to generate or redesign only parts of a web. The use of a generative pre-trained transformer strategy with a decoder-only makeup provides further evidence for the universal applicability of these models beyond language (such as done in GPT-3/4, etc.), to include complex physical systems[78,85,86].

The results reported in this paper provide a platform to explore structural-property relationships of spider webs and other graph architectures, including structural performance optimization, and facilitates learning web design principles from natural intelligence. Future works include computational works and analysis for spider webs using additional, augmented datasets (e.g., more original web graphs), improving the *de novo* design algorithm of spider webs for practical applications, as well as mechanical tests incorporating 3D printing techniques for synthetic design assessment. In addition, the developed deep generative models featuring high generalizability and learning capacity for large-scaled graph generation for spider webs, which allow graph conditioning, both categorial and continuous label learning, and domain-agnostic data input, have the potential to be applied beyond spider webs to a broad range of bio-inspired designs that exhibit similar hierarchical properties, thus providing abundant opportunities for research.

## 4. Materials and Methods

### Dataset construction and conditioning parameters $C$

Building on the hierarchical structure of spider webs, in order to capture invariant construction principles and the high degree of heterogeneity (**Figure 1b-c, Figure S2**), we use inductive representative sampling as suggested in[75] to generate smaller subgraphs and improve the construction of local structures (schematically demonstrated in **Figure 1d-f**). Permutation invariance is considered in this method, since the sampling process is independent of the selection order of central nodes, and there could be overlapping sections among subgraphs that consists of different node orderings[75]. The analysis and calculation are based on the original spider web data (**Figure 1d**), which are presented in SI units[42].

This strategy is motivated by our desire to learn the construction principles by which spiders build webs. Each inductively generated web sample graph $\mathcal{G}_i$, starting at one node in each of the large webs, is unique. We use 90% of the data for training, and 10% to generate and validate generated samples (since each of the samples is unique, the testing conducted and shown in **Figures 3d-f** are significant in terms of predictive generalization capacity).

We calculate seven geometric properties to describe characteristics of each web sample (statistics of the distribution shown in **Figure S4**). The parameters include the number of nodes in the graph, $N_N$, and the mean of the edge lengths:

$$\bar{l} = \frac{1}{N_E} \sum_{i=1..N_E} \sqrt{dx_i^2 + dy_i^2 + dz_i^2} \tag{1}$$

where $d(x, y, z)_i$ is the length of the $x, y$ and $z$-component of edge $i$, and $N_E$ is the total number of edges in the graph. Similarly, we calculate the mean of the $x, y$ and $z$ components of the edge lengths:

$$\overline{dx} = \frac{1}{N_E} \sum_{i=1..N} dx_i, \quad \overline{dy} = \frac{1}{N_E} \sum_{i=1..N} dy_i \quad \overline{dz} = \frac{1}{N_E} \sum_{i=1..N} dz_i \tag{2}$$

to capture residual directional features of the edges. The node degree is defined as

$$\zeta = \frac{N_E}{N_N} \tag{3}$$

These features are summarized in a 7-dimensional feature vector $C_i$ (which is used for graph conditioning in the generative process), for each sample $i$:

$$C_i = [\bar{l}, \overline{dx}, \overline{dy}, \overline{dz}, N_N, N_E, \zeta] \tag{4}$$

The maximum length of nodes considered is denoted as $N$, whereas $N \geq \max_i(N_{N,i})$, considering all samples $i$ in the dataset. For the cases considered in this paper, $N = 64$. The conditioning features are normalized to lay between -1…1, whereas each feature within $C_i$ is treated separately. The resulting distributions are shown in **Figure S4**.

### Model 1: Analog diffusion model with sparse neighbor representation

In this approach, we describe graph representations as defined in **Figure S3.** Since we use a sparse representation, the input structure consists of a list of embeddings of length $N$, padded with zeros for graph samples of length less than $N$. The representation $z$ is defined by a sequence of elements $z_j$ for each node (where $j = 1..N$), defined as

$$z_j = [x_j, y_j, z_j, \mathcal{N}_{1,j}, \mathcal{N}_{2,j}, .., \mathcal{N}_{\text{Neigh}_{\max},j}]^T \tag{5}$$

where $\text{Neigh}_{max}$ denotes the maximum number of neighbors considered, and $\mathcal{N}_{k,j}$ refers to the node number of neighbor $k$ of node $j$ (if a node has fewer neighbors than $\text{Neigh}_{max}$, we pad the excess entries with 0). We determined that the spider web graphs considered have no more than six neighbors at each node, hence $\text{Neigh}_{max}=6$. With $z_j$ defining each row, the full matrix representation is defined as

$$z = [z_1, z_2, \ldots, z_N] \quad (6)$$

It is noted that $z$ is a matrix of size $[3+\text{Neigh}_{max}, N]$.

The coordinate components $x_j, y_j, z_j$ in $z_j$ are normalized to be between -1…1 (we normalize the coordinate components with a single set of normalizing parameters found across all three components). Reflecting the use of an analog representation of the discrete neighbor lists, the neighbor components are normalized in the same way, also to yield values between -1…1. This allows us to formulate the problem as a continuous denoising problem, following the concept of analog representations of discrete variables. During sampling and subsequent graph construction, all predicted features $z$ are unscaled, and the neighbor descriptors $\mathcal{N}_{k,j}$ are rounded to the nearest integer number to yield a discrete representation. **Table S2** provides a summary of the parameters used to construct the neural network architecture.

*Denoising process*

**Figure 2d** visualizes the denoising process, where the top defines a Markov chain operator q that adds Gaussian noise step-by-step (according to a defined noise schedule that defines how much noise $\varepsilon_i$ is added at each step $i$), translating the physical solution (either, the stress-deformation response or the encoded image) $z_0$ (left) into pure noise, $z_F$ (right) following:

$$z_{i+1} = \text{q}(z_i) \quad (7)$$

The deep neural network is trained to *reverse* this process. This is done by identifying an operator p that maximizes the likelihood of the training data. Once trained, it provides a means to translate noise to solutions and thereby realizing the transition illustrated in **Figure 2a** (noise to solution), in a step-by-step fashion as indicated in the lower row of **Figure 2d**:

$$z_{i-1} = \text{p}(z_i) \quad (8)$$

We use the L2 distance to define the loss, measuring the error of the actual added noise $\varepsilon_i$ and the predicted added noise $\varepsilon_i'$. Hence, the trained diffusion model can predict the added noise. Knowing this quantity then allows us to realize a numerical solution to the denoising problem and hence be used to generate the next iteration of the denoised sequence:

$$z_{i-1} = z_i - \varepsilon_i' \quad (9)$$

In eq. (9), the sequence $z_i$ at step $i$ is transformed by removing the noise $\varepsilon_i'$. This process is performed iteratively; whereas the neural network predicts, given the current state $z_i$, the noise to be removed at a given time step $t_i$ in the denoising process (see **Figure 2d**). The conditioning signal $C_i$ is encoded using additive Fourier positional encoding following the approach suggested in[87]. For cross-attention as shown in **Figure 2c**, the conditioning signal is concatenated with the time signal so that $C_{i, \text{total}} = [C_i, t_i]$. This allows us to train a single neural network to learn denoising steps, while considering the conditioning and the time step $t_i$ in the reverse process. The U-net architecture features 1D convolutional layers mixed with self-/cross-attention layers (see eqs. (13-14) and associated discussion for more details). The convolutional layers capture hierarchical patterns, and the attention layers learn long-range relationships between patterns. This type of architecture can effectively capture the complex mechanisms in the process of spider web construction.

We use the noise schedule, sampling and training processes proposed in proposed in[88] since it yields computationally efficient sampling strategies, obtaining results within 96 denoising steps. **Table 1** provides additional details about the model architecture parameters.

The functioning of both models in forward and inverse directions is schematically shown in **Figure 2d**.

**Model 2: Diffusion model with full neighbor matrix representation**

The model is constructed similarly to Model 1, except for the representation $z$. Here, $z$ captures node positions and a full adjacency matrix, where the sequence of elements $z_j$ for each node, defined as

$$z_j = [x_j, y_j, z_j, \mathcal{N}_{1,j}, \mathcal{N}_{2,j}, \ldots, N]^T \tag{10}$$

where $N$ denotes the maximum number of nodes considered, and $\mathcal{N}_{k,j}$ denotes whether node $j$ has a neighbor $k$:

$$\mathcal{N}_{k,j} = \begin{cases} 1: \text{if node } k \text{ is a neigbor of node } j \\ -1: \text{else} \end{cases} \tag{11}$$

With $z_j$ defining each row, the full matrix representation is defined as

$$z = [z_1, z_2, \ldots, z_N] \tag{12}$$

It is noted that $z$ is a matrix of size $[3+N, N]$. Just like in the previous model, the diffusion model is trained to learn a conditional denoising process, but now on these larger representations. **Table S3** provides a summary of the parameters used to construct the neural network architecture.

**Model 3: Autoregressive transformer architecture with full adjacency matrix representation**

**Figure 2e-g** depicts a summary of the transformer architecture, representing an autoregressive decoder-only architecture that produces solutions iteratively from a start token during inference and using cross-attention with the conditioning features $C'_i$, where $C'_i$ is the output of a feed-forward layer that expands the dimension from $[7, 1]$ to $[7, d_C]$. The key mathematical operation is the masked attention mechanism[78,89], defined as

$$\text{Attention}(Q, K, V; M) = \text{softmax}\left(\frac{QK^T + M}{\sqrt{d_k}}\right) V \tag{13}$$

The attention calculation is implemented in multi-headed form by using parallelly stacked attention layers. Instead of only computing the attention once, in the multi-head strategy we divide the input into segments (in the dimension of the embedding) and then computes the scaled dot-product attention over each segment in parallel, allowing the model to jointly attend to information from different representation subspaces at different positions:

$$\text{MultiHead}(Q, K, V) = \text{Concat}(\text{head}_1, \ldots, \text{head}_h)W^O \tag{14a}$$
$$\text{head}_i = \text{Attention}(QW_i^Q, KW_i^K, VW_i^V) \tag{14b}$$

where the projections are parameter matrices $W_i^Q \in \mathbb{R}^{d_{model} \times d_q}$, $W_i^K \in \mathbb{R}^{d_{model} \times d_k}$, $W_i^V \in \mathbb{R}^{d_{model} \times d_v}$ and $W^O \in \mathbb{R}^{hd_v \times d_{model}}$. In self-attention, all $Q, K, V$ come from either input or output embeddings (or other sources) only, cross-attention calculations here are performed with $Q$ from the conditional embeddings and $K, V$ from the encoded desired graph properties, computed from the input embedding via a fully connected feed forward network from the 7-dimensional feature vector $C_i$. During training, the target represents web samples, conditioned by the associated graph properties $C_i$ whereas the model is trained to predict the output. Causal masking using a triangular masking matrix $M$ is used in the self-attention step so that the model can only attend to tokens to the left (*i.e.*, previous tokens).

As shown in **Figure 2g**, a start token $\mathcal{T}$ is added at the beginning of the graph descriptor sequence, so that

$$z = [\mathcal{T}, z_1, z_2, \ldots, z_N] \tag{15}$$

During generation, the start token $\mathcal{T}$ is first fed into the model and the output is predicted from it. During sampling iterations, this process is repeated until the full graph is produced, as shown schematically in **Figure 2f-g.** The graph descriptor $z$ is identical to the definition above, since positions with a full distance matrix are considered. **Table S4** provides a summary of the parameters used to construct the transformer neural network architecture.

**Training process and other hyperparameters**

All code is developed in PyTorch, for further details see[90]. All machine learning training is performed using an Adam optimizer[91], with a learning rate of 0.0002.

*De novo* **graph construction**

To construct larger *de novo* graphs, we use the three models described above to generate smaller graph samples $\mathcal{G}_i$, and then combine them to form larger graphs $\mathcal{G}$, where

$$\mathcal{G} = f[\mathcal{G}_1, \mathcal{G}_2, \ldots, \mathcal{G}_n] \tag{16}$$

whereas each graph is defined by $z$, a matrix of size $[3+N, N]$ for *de novo* graph construction, we operate in representations of graphs with full adjacency matrix; and predicted graphs – if not in that format (such as in Model 1) – are transformed into such a representation before the operations are conducted.

The function $f$ denotes how individual graphs $\mathcal{G}_i$ are assembled. In the examples presented in the paper, assembly is done iteratively where two graphs $\mathcal{G}_i$ and $\mathcal{G}_{i+1}$ are combined to form a new graph $\bar{\mathcal{G}}_{i+2}$. This process is repeated iteratively to form a new graph from the earlier ones

$$(\mathcal{G}_i, \mathcal{G}_{i+1}) \rightarrow \bar{\mathcal{G}}_{i+2} \tag{17}$$

Reflecting various choices of $f$ different strategies can be implemented to connect two graphs $\mathcal{G}_i$ and $\mathcal{G}_{i+1}$. If the same graph is used, symmetric shuffling of the graph representation $z$ as depicted in **Figure 4a** helps to randomize the connectivity of the new graph without changing the representation of the graph itself since it is invariant to such operations (note, in **Figure 4a** we only show the adjacency matrix, not the positions, but these are also shuffled in the same way as the connectivity to yield structural invariance). To stack two graphs, we overlay the two neighbor matrices as shown in **Figure 4b**, to form a larger graph. Since all graphs generated by the models during samples are centered around (0, 0, 0), to realize spatially distributed graphs, the coordinates $x, y, z$ of the second graph $\mathcal{G}_{i+1}$ are iteratively shifted by *dx, dy* and *dz*, according to some function

$$f_c(x, y, z) = [x + \Delta f_{c,x}, y + \Delta f_{c,y}, z + \Delta f_{c,z}] \tag{18}$$

Generally, the functions $\Delta f_{c,x/y/z}$ depend on the iteration number in the graph generation process. Examples of coordinates produced from these calculations are shown in **Figure 4c/d**, center panel. Coordinates in the overlapping region (marking in **Figure 4b**) are either averaged from $\mathcal{G}_i$ and $\mathcal{G}_{i+1}$ or taken from the second graph, $\mathcal{G}_{i+1}$.

In the cases where we generate helices, the vector function $f_c(x, y, z)(i)$ is defined as:

$$f_{c,x} = R\cos(i\Delta\phi) \tag{19a}$$

$$f_{c,y} = R\sin(i\Delta\phi) \tag{19b}$$

$$f_{c,z} = it \tag{19c}$$

where $\Delta\phi$ denotes the angular increment in each step, $R$ the radius of the generated helix, and $t$ the slope in the $z$ direction.

In examples using parametric curves (**Figure 5a**), the vector function $f_c(x, y, z)(i)$ defines a closed curve in cartesian space through the set of parametric equations below:

$$f_{c,x} = (A + \cos(Bit))(\cos(it)) \tag{20a}$$

$$f_{c,y} = (A + \cos(Bit))(\sin(it)) \tag{20b}$$

$$f_{c,z} = \sin(Bit) \tag{20c}$$

Where $A$ and $B$ are constants equal to 2 and 1.5 respectively and the term $it$ traverses a domain from 0 to 12.57, with a step size of $t = 0.05$ (via 251 steps).

In the samples defined along curve paths, the parametric functions are defined by the system of parametric equations above. Equations were solved at timesteps along the described domain of t to generate coordinates in cartesian space along which a curve was defined by line segments connecting each point.

**Additive manufacturing and mechanical testing**

Samples of webs were 3D printed for the purposes of visualization and mechanical testing using a Stratasys Objet 500 Connex V3 polyjet 3D printer (Stratasys Inc., Revohot Israel) (**Figures 5c-d and 6b**). Stereolithography (STL) file formats were generated from web geometries using the marching cubes algorithm[92] to create a mesh with a constant radius of 0.4 mm along the edges of the spiderweb graph (several STL files are provided in **Supplementary Information**). This mesh was converted into a signed distance field (SDF) volume and smoothed in order to create more consistent transitions at nodes. This volume was then reconverted into a triangulated mesh and exported for additive manufacturing. Prints were manufactured using polyjet printing with the Stratasys RGD 450 resin and dissolvable support (SUP706B) and cleaned in a base bath prior to display and testing.

The helix models for tensile testing were scaled isometrically so that their maximum dimension was equal to 10 cm prior to meshing and manufacturing. The 3D model was blended with block geometries at each end to provide an area to attach to the grips on the mechanical testing apparatus. To prevent breakages near the grips, the 3D model was converted to a volume and smoothed before remeshing, creating denser solid regions blended into the grip regions. Tensile testing was performed using an Instron Universal Testing Machine (Instron, Norwood, MA) with a 5 kN load cell. Samples were strained at a rate of 0.5 mm/min until failure. **Movies M1-M3** show the three tensile tests, with a speedup of 50x.

**Authors' Contributions**

M.J.B. developed the overall concept and the algorithm, designed the ML model, developed the codes, oversaw the work, and drafted the paper. W.L. and M.J.B analyzed the data and edited and wrote the manuscript. W. L. and N.A.L conducted mechanical tests. N.A.L. contributed to the analysis of the web statistics, visualization, parametric and differential equations, and additive manufacturing of generated web samples.

**Acknowledgements**

We acknowledge support by NIH (5R01AR077793-03), the Office of Naval Research (N000141612333 and N000141912375), AFOSR-MURI (FA9550-15-1-0514) and the Army Research Office (W911NF1920098). Related support from the MIT-IBM Watson AI Lab, MIT Quest, and Google Cloud Computing, is acknowledged.

**Data and Code Availability**

Codes and data are available via GitHub: https://github.com/lamm-mit/GraphGeneration/.

**Supplementary Information**

- Supplementary text, references, and Supplementary Figures
- STL Files for the three helical webs tested under tension (H1.stl, H2.stl, H3.stl) and two parametric bounding curve webs (boundary_1.stl, boundary_2.stl): https://www.dropbox.com/s/fezzk110gh5udws/STL%20Files.zip?dl=0
- Video data showing the tensile tests conducted (M1.mp4, M2.mp4, M3.mp4), as well as an overview of 3D renderings of three of the designs (M4.mp4).

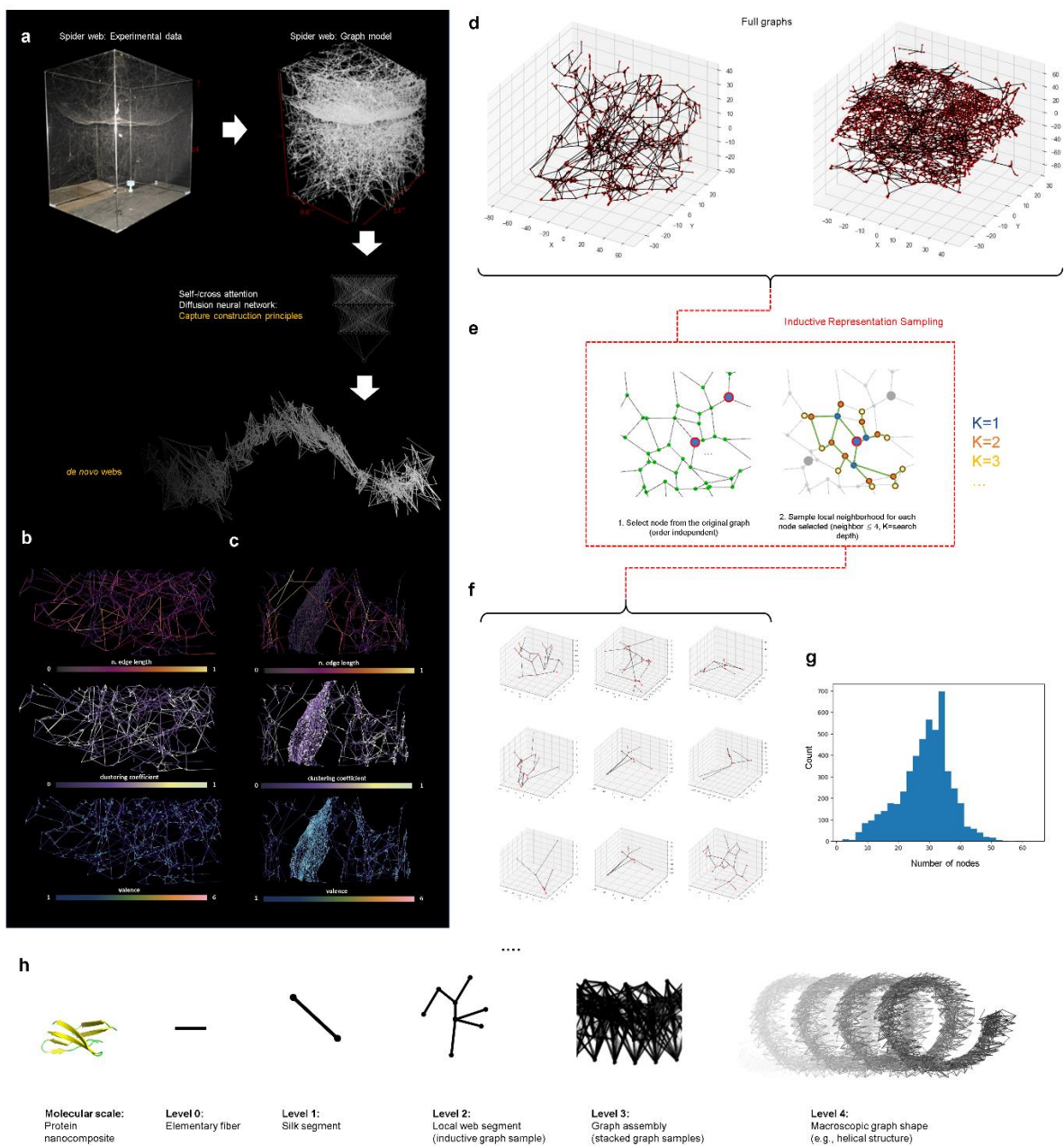

**Figure 1.** Overview of the work reported, including details on graph representation strategies. Panel **a:** Summary of the overall approach reported in this paper. Based on a spider web scanned using laser tomography[24–26,29,32], we develop a graph model, which is then modeled, and transformed into, new bio-inspired, synthetic structures. As shown in panels **b and c** for two webs, the structures feature a high degree of heterogeneity (via coloring we show the variation of edge length, clustering coefficient [72], and valence (number of neighbors for each node) (further analysis see **Figure S2**). Panels **d-f**: development of web construction representations using inductive sampling[75]. Samples of large webs (**d**) are developed into a large training set (**f**) via inductive sampling, considering up to fourth nearest neighbors (this process is repeated for all nodes in each of the large webs shown in panel **d**). Panel **g** shows the length distribution of the resulting graphs. Panel **h** shows the hierarchical structuring used here, by breaking down a complex graph (Level 4) into a local graph assembly (Level 3), a local web segment as defined by inductive sampling (Level 2), individual silk segments (Level 1, defined by a finite length fiber with a start and end point), and the elementary fiber (Level 0). At the ultrascale, silk consists of a molecular-level nanocomposite, which is not considered here.

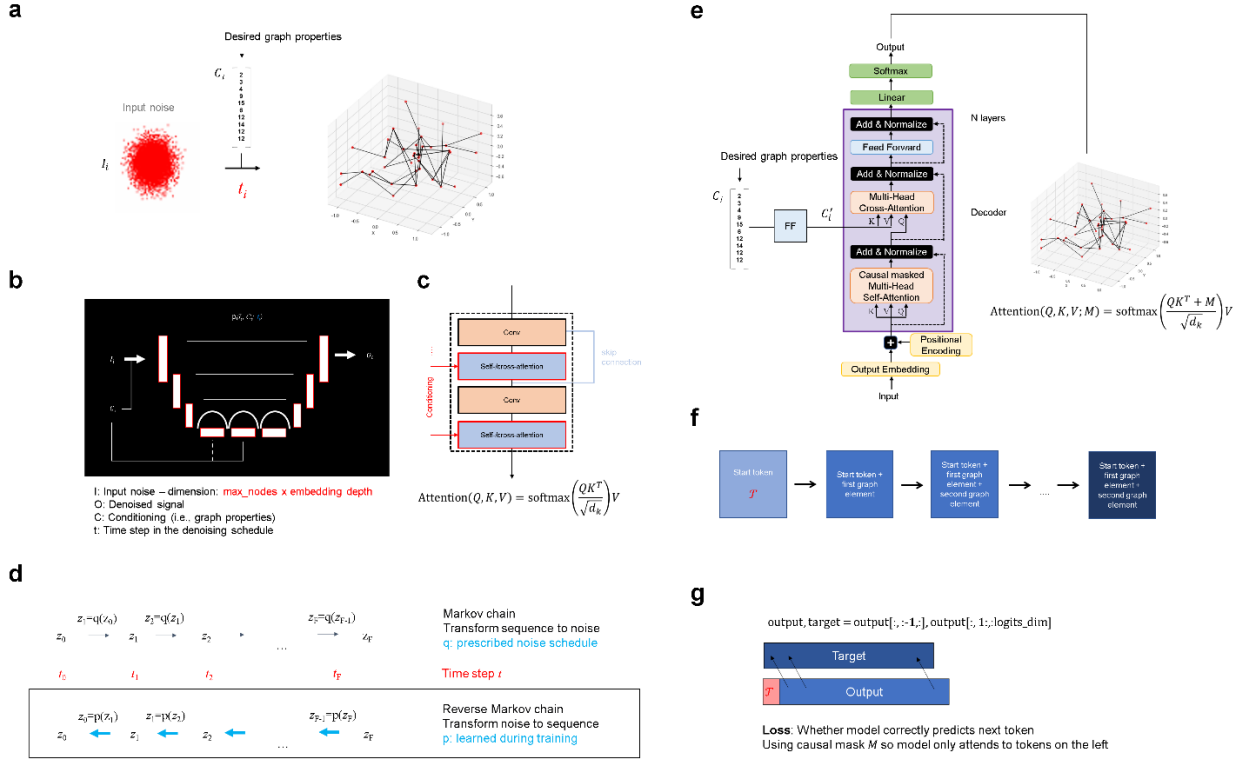

**Figure 2.** Deep learning architectures used in this work, featuring attention-based diffusion models and autoregressive transformer architectures, also featuring the attention mechanism. Panel **a-d**: Summary of the diffusion algorithm used to translate noise to graphs (**a**). Panel **b** shows the U-net architecture used here, along with details on the combined cross-attention architecture to realize conditional denoising (panel **c**) with respect to both, the conditioning variables *C* and time step *t*. Panel **d** shows the forward Markov and trained reverse Markov chain that is learned by the U-net architecture to predict a conditioned denoising process, whereas a multi-headed attention mechanism is used to learn long-range relationships (eqs. (13-14)). Panel (**e**): Summary of the transformer architecture, representing an autoregressive architecture (**e-f**) that produces solutions iteratively from a start token during inference. During training (**g**), the target represents web samples, conditioned by the associated graph properties. The model is trained to predict the proper output. A start token is added at the beginning of the graph descriptor sequence. During generation, the start token is first fed into the model and the output is predicted from it. This process is repeated until the full graph is produced.

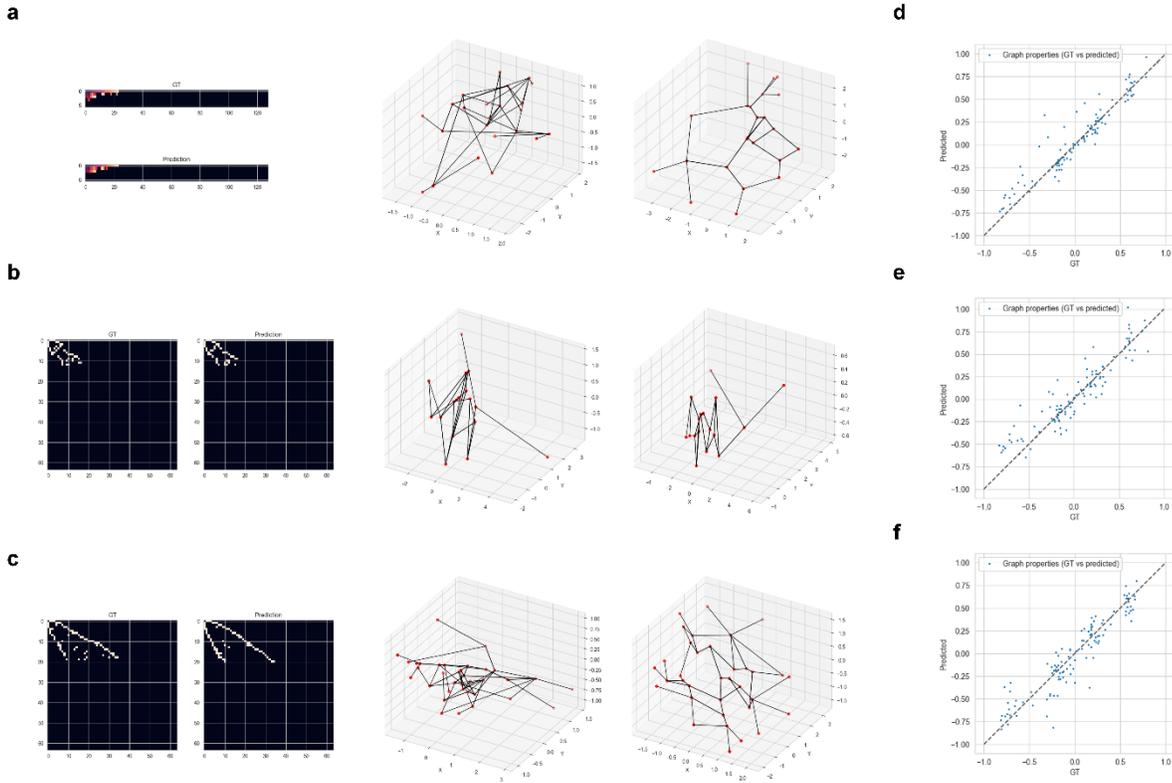

**Figure 3.** Sampling results using all three models, including validation experiments to confirm that the conditioning produces graphs with required properties**.** Sample generated web sections using **a**, the sparse analog diffusion model, **b**, the diffusion model with full neighbor representation, and **c**, the autoregressive transformer architecture. Panels **d-f** show results of testing the correlation of conditioning variables (x-axis, labeled GT, over the entire test set $\{C_i\}_{\text{test,GT}}$) with measured properties of the generated graphs in a test set of webs not used for training (y-axis, labeled Predicted, $\{C_i\}_{\text{predicted}}$). The analysis is conducted over the entire distribution of samples as depicted in **Figure S4.** The results are shown for the three models: **d,** a sparse analog diffusion model. **e**, analog diffusion model (non-sparse), and **f,** autoregressive transformer model (non-sparse). The R2 values (from top to bottom) are identified as 0.89, 0.84, 0.83. The sparse analog diffusion model (panel **d**, and shown in detail in **Figure 2a-d**) features the best performance. Since each of the samples is unique, the testing conducted and shown in **Figures 3d-f** are significant in terms of predictive generalization capacity.

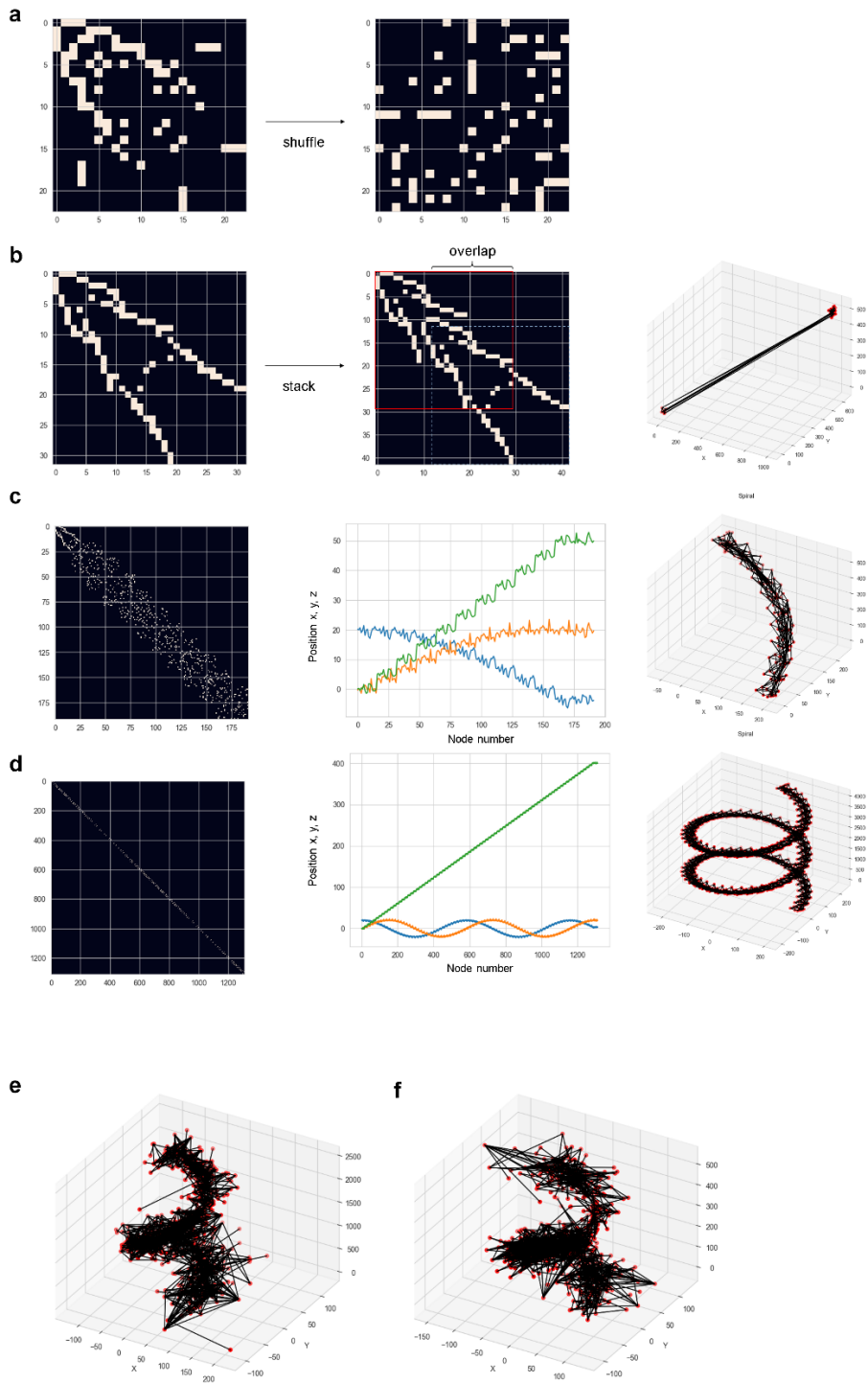

**Figure 4.** Algorithm utilized to generate larger-scale *de novo* graphs, based on the concept of stacking graph samples into larger graph assemblies. Panel **a** shows a shuffling algorithm by which the node positions and adjacency matrix are shuffled symmetrically. While this operation is invariant to the representation of the graph, it allows us to obtain more diverse stacking results and hence helps to randomize the inductively sampled graphs and achieve greater diversity of how nodes of stacked graphs are connected. Panel **b** shows an example of how graphs are stacked in *z* space, forming a larger graph (here, shown for two identical graphs as the basis). Coordinates in the overlapping region are either averaged or taken from the second graph; and coordinates of identical graphs are shifted by *dx, dy* and *dz* (in the example shown on the right, extreme choices of displacements are used to visually represent the new connections, and new graph, formed). Panel **c** shows an example of a larger-scale stacking, repeated multiple times, and defining *dx, dy* and *dz* (to form a helix. The resulting graph is shown on the right. Panel **d** shows another example, here without shuffling, forming a larger helical graph. The center graphs in panels **c** and **d** show the coordinates of the nodes, over node numbers. Panels **e** and **f** show examples of continuously sampled graphs, producing complex architectures. In the two examples, we use the autoregressive transformer model (**e**) and the analog diffusion model (**f**) to realize graphs. At each stacking step, a new graph is sampled, conditioned based on a randomly generated conditioning vector $c$.

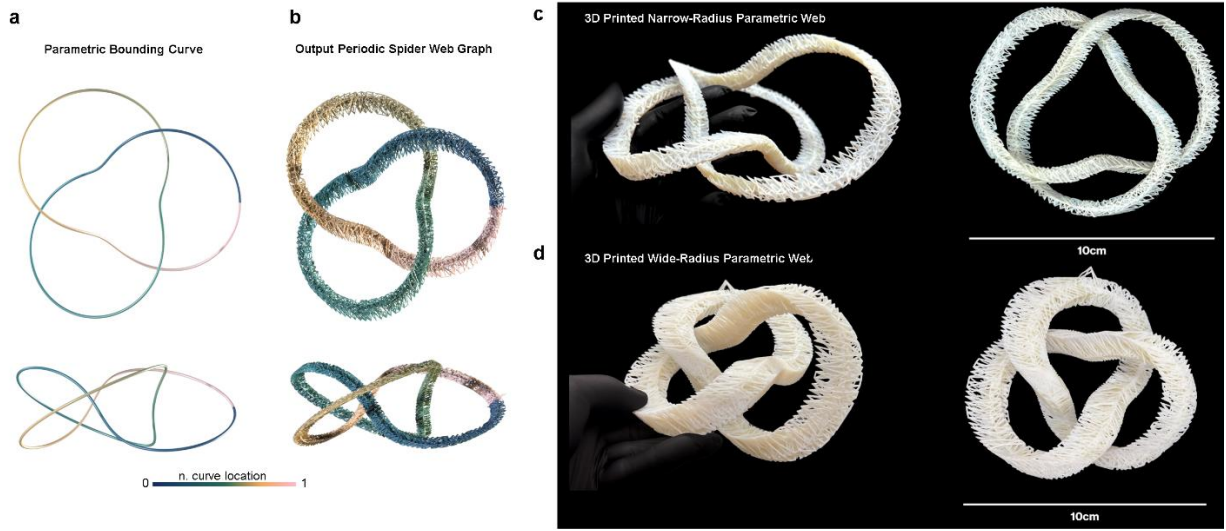

**Figure 5.** Complex web formations were generated along paths defined by bounding curves. A system of parametric equations in cartesian space defined a closed bounding curve (**a**) that served as an input geometry for the non-sparse analog diffusion model, outputting a lattice-like web formed along the bounding curve (**b**). Panels **c** and **d** show webs generated along parametric bounding curves with a strut radius of 0.4 mm and 3D were printed via polyjet printing. Variations of thin (**c**) and thick (**d**) path radii were used to generate distinct geometries.

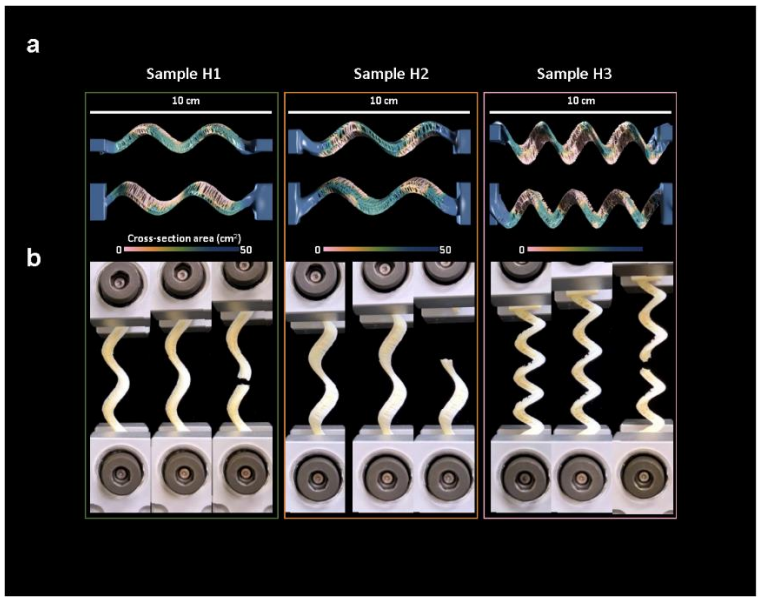

**Figure 6.** Helical web geometries are meshed and merged with blocks to provide a region that could be attached to a mechanical testing apparatus. Cross-sectional areas are calculated in slices perpendicular to the force vector of the mechanical test and are color coded in panel **a**. Slices with multiple distinct regions had their areas calculated separately. The extension of samples during the tensile test are shown in panel **b**, with output data displayed in panel **c**. Recorded testing videos are provided in **Supplementary Information**. Samples H1, H2 and H3 vary in coil number (2:2.5:4), and sample H3 is designed with a wider-radius web compared with H1 and H2. The strut breaking events, mechanisms of failure, are observed near the middle along each helix design.

# Modeling and design of heterogeneous hierarchical bioinspired spider web structures using generative deep learning and additive manufacturing

Wei Lu[1,2], Nic A. Lee[1,3], Markus J. Buehler[1,2,4*]

[1] Laboratory for Atomistic and Molecular Mechanics (LAMM), Massachusetts Institute of Technology, 77 Massachusetts Ave., Cambridge, MA 02139, USA

[2] Department of Civil and Environmental Engineering, Massachusetts Institute of Technology, 77 Massachusetts Ave., Cambridge, MA 02139, USA

[3] MIT Media Lab, Massachusetts Institute of Technology, 77 Massachusetts Ave., Cambridge, MA 02139, USA

[4] Center for Computational Science and Engineering, Schwarzman College of Computing, Massachusetts Institute of Technology, 77 Massachusetts Ave., Cambridge, MA 02139, USA

# SUPPLEMENTARY MATERIALS

**Legends for supporting video files**

- **Movie M1:** Tensile deformation experiment of H1, 50x speedup.
- **Movie M2:** Tensile deformation experiment of H2 50x speedup.
- **Movie M3:** Tensile deformation experiment of H3, 50x speedup.
- **Movie M4:** Rendering of various 3D geometry files generated in this paper.

## S1. Overview of graph generation models

Different types of deep graph generation models have been proposed, each with distinct characteristics and learning capabilities. Typically implemented model architectures are variational autoencoders (VAEs), generative adversarial networks (GANs), autoregressive models (ARs), and diffusion models[44,60]. The illustration for each architecture is shown in **Figure S1**, and existing generators are listed in **Table S1** in the supplementary material. Common model structures are discussed below.

- In VAEs, the input graph is encoded into a lower-dimensional latent space, and the distribution of essential features is learned and output from the decoder, then sampled during graph generation[46,61]. Generation capacity and graph diversity are the main limitations for basic VAE graph generators, for example, GraphVAE[62]. Therefore, several improved models have been developed though they are still largely limited to small-sized graph generation. Graphite[63] applies an iterative graph refinement strategy to improve the computational efficiency, while DGVAE[64] incorporates balanced graph cut, CGVAE[65] adapts graph linearization for better generation quality, and JT-VAE[66] utilizes hierarchical molecule representations, and uses substructures as building blocks.
- In GANs, a generator network is trained to generate realistic graphs with random seeds as inputs. Feeding graph data, the discriminator network outputs probability to distinguish real from generated graphs. During training, both the generation and detection ability of the generator and discriminator are enhanced, and higher-quality graphs are designed[67]. GANs still suffer from limitations when it comes to graphs with large sizes and high complexity. GAN-based graph generation models include: NetGAN[68] which combines random walks for graph generation; CondGen[69] enables flexible semantic graph conditioning; LGGAN[70] generates labeled graphs and supports node labels for improved feature capturing; and TSGG-GAN[71] explores interrelationships between time series data.
- In ARs, graphs are sequentially generated with modeling probability distributions at each time step based on previously generated sections. The sequence-independent process requires relatively more computational costs but allows for larger-sized and more coherent graph designs than one-shot generations[47]. GraphRNN[72] is one of the most commonly used autoregressive models for graph generation, with edge sequences constructed based on node sequences. GraphRNN increases the generalizability, and provides innovative evaluation metrics, but its incapability of learning nodal features is one major limitation. Compared with GraphRNN, GraphGEN[73] increases the scalability of generation, enables the training of features and labels for nodes and edges, and provides comprehensive evaluation metrics for structural metrics, entity labels, graph property, generation similarity and novelty. Besides, GRAN[74] and BiGG[75] enhance training efficiency respectively through generating blocks of graphs instead, and incorporating the binary tree.
- Diffusion models are a type of iterative generative architecture[76] that can be adapted for graph generation. An important trade-off when using diffusion models is between the learning capacity and computation cost. In its diffusion process, random Gaussian noise is added to input each time step, through a prescribed algorithm, via a Markov chain. Then in the denoising process, added noise is learned from training to successively denoise the data, to yield an inversed Markov chain. The output sequence is then translated to graph data format[77]. For specific model applications, the EDP-GNN[78] model achieves permutation invariance through score matching. Based on EDP-GNN, GraphGDP[59] applies a position-enhanced score network to include more topology information for improved generation, especially for larger graphs. NVDiff[79] combines VAE to reduce the diffusion dimension and foster sampling efficiency; GSDM[80] argues the full adjacency matrix training impedes the learning and sampling speed, and introduces a spectrum-based algorithm; and DiGress[81] adds conditions of structural and spectral graph features to the diffusion process to improve generation.

There exist many other deep learning-based generators (as listed in **Table S1**), but most have been applied explicitly for molecular data with relatively smaller scales and specific types of representations. These models are typically trained only on atom-level labels (such as position, atom type, and properties). The edge connection is then reconstructed by identifying the bond type according to atomic analysis (e.g., electronegativity)[58,82]. In addition, the nature of generic graph-structured data leads to permutation invariance and accordingly increases time complexity. Multiple edge representations are possible for a single graph due to node orderings, though the same probability is anticipated for data distribution. Specifically, permutation invariance implies $p(A^{\pi_1}) = p(A^{\pi_2}) \; \forall \; \pi_1 \neq \pi_2$, where $A$ denotes the edge representation and $\pi$ represents the node ordering. The following approaches are investigated in existing models, yet do not fully address the issue in general[78]: data augmentation through random permutation[44]; inexact graph matching[62]; modified graph linearization[65]; random breadth-first search[72,73]; pre-defined node ordering selection[74]; node embeddings to single graph embedding translation[69]; and score matching[59,78,83].

**Supplementary figures**

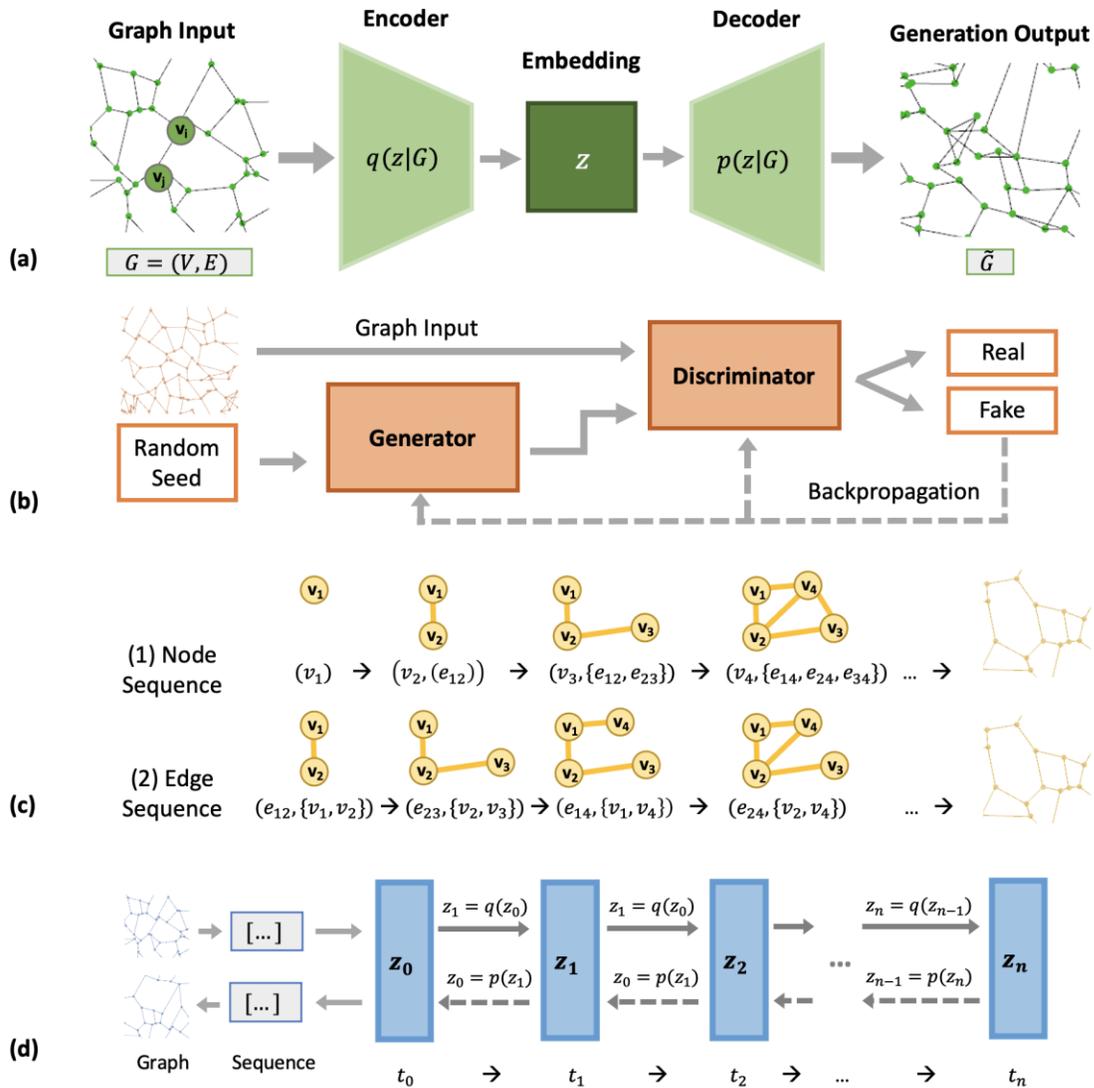

**Figure S1.** Common model architectures for deep graph generators. An undirected graph is defined as $G = (V, E)$, where $V = \{v_1, v_2, ..., v_n\}$ and $E = \{e_{ij} = (v_i, v_j) | v_i, v_j \in V\}$ represent node and edge sets respectively. **(a)** Variational autoencoder (VAEs): the encoder $q(z|G)$ convert input graph $G$ into lower-dimensional embedding $z$, then decoder $p(G|z)$ outputs probability and new graph $\tilde{G}$ is sampled[1,2]; **(b)** Generative adversarial network (GANs): the generator is trained to generate realistic graphs with random seeds as inputs, and the discriminator network outputs probability to classify real and fake graphs to improve the generation[3]; **(c)** Autoregressive models (ARs): both node or edge-based sequence generation at each time step is conditioned on previously generated sections[4]; **(d)** Diffusion models: in the diffusion process, random Gaussian noise is added to input $z_t$ every time step $t$ through a prescribed schedule $q(z_{t-1})$, and in the denoising process, the noise is converted back to sequence through the inversed algorithm $p(z_{t+1})$,, then translated to graph data[5].

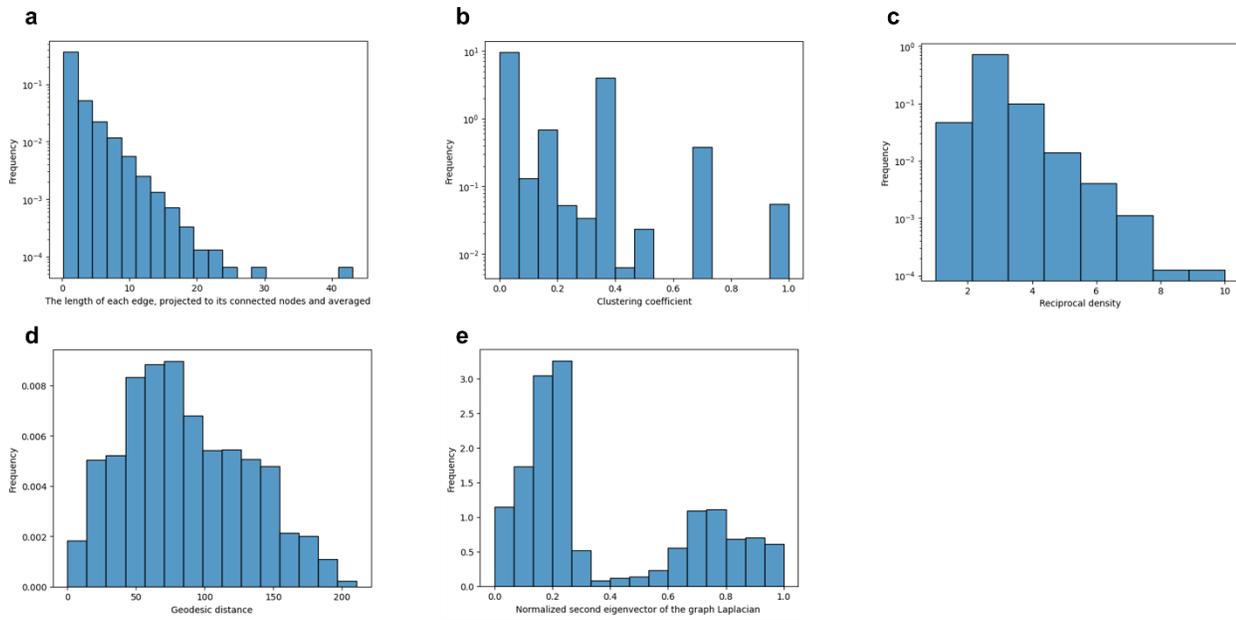

**Figure S2:** Statistical analysis of key geometric graph properties of the spider webs, complementing the visual analysis shown in **Figure 1b-c**. Panel **a** shows the length of each edge, projected to its connected nodes and averaged. Panel **b** shows the clustering coefficient, and **c** the reciprocal density of neighbors (we calculate this by getting the distance to each neighbor in the graph and dividing it by the number of neighbors; this means that low values indicate high density). Additional analyses are shown in panels **d-e**. Panel **d** depicts the geodesic distance for each node from an arbitrary point on the web's edge (also provides an idea of signal transduction). The number per node is not important, but its variance gives a metric connectivity and efficiency of travel across the web (high variance means more localized regions, low variance means a consistent level of density and connectivity). Panel **e** shows the normalized second eigenvector of the graph Laplacian projected on a per node basis. More similar values indicate "neighborhoods" or anatomy (perhaps indicating the way a spider would travel through it). High variance indicates more isolated regions of anatomy. All data analysis is done in SI units.

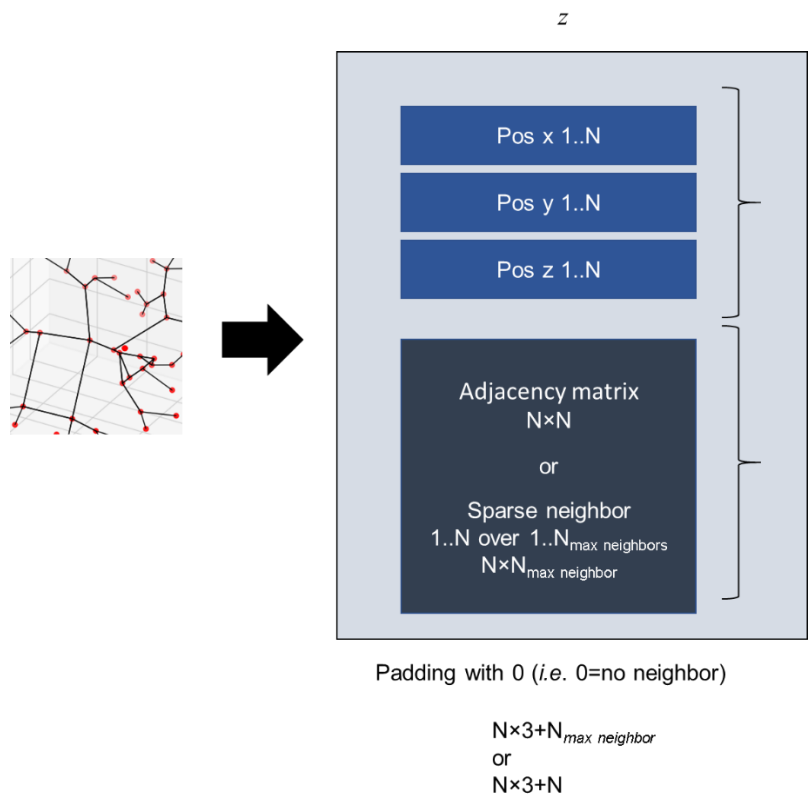

**Figure S3.** How graphs are represented in the deep learning models developed in this paper, referred to as the $z$ space. Each inductively generated sample is encoded as described in the figure, using either a full adjacency matrix or a sparse neighbor representation, where a maximum number of neighbors is denoted.

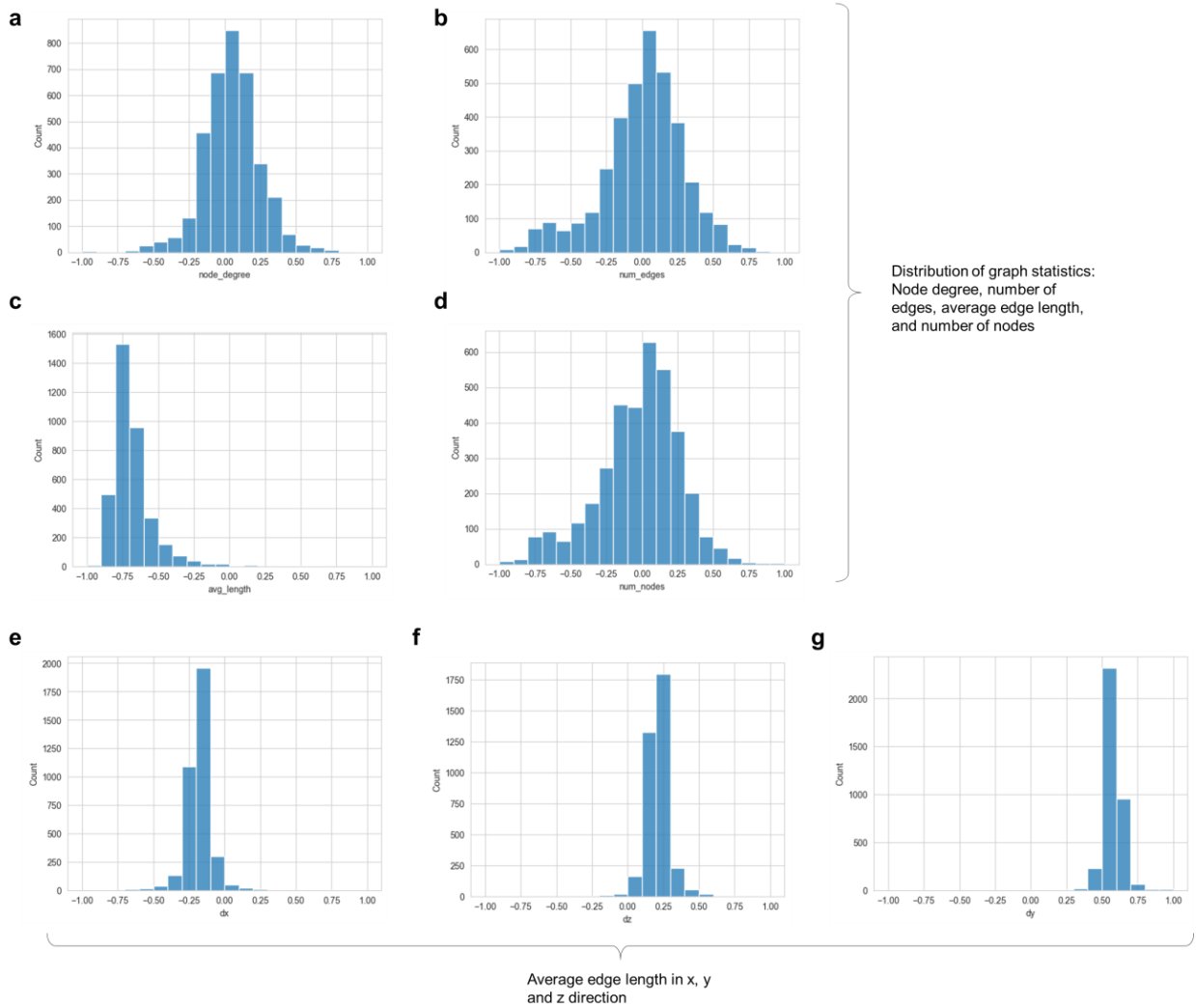

**Figure S4.** Properties used to condition the generative models, featuring a total of 7 geometric properties as defined in equations (1)-(4), captured in $C$. The plots show histograms of each of the variables over the entire set of $C_i$ extracted from the original experimentally acquired spider webs.

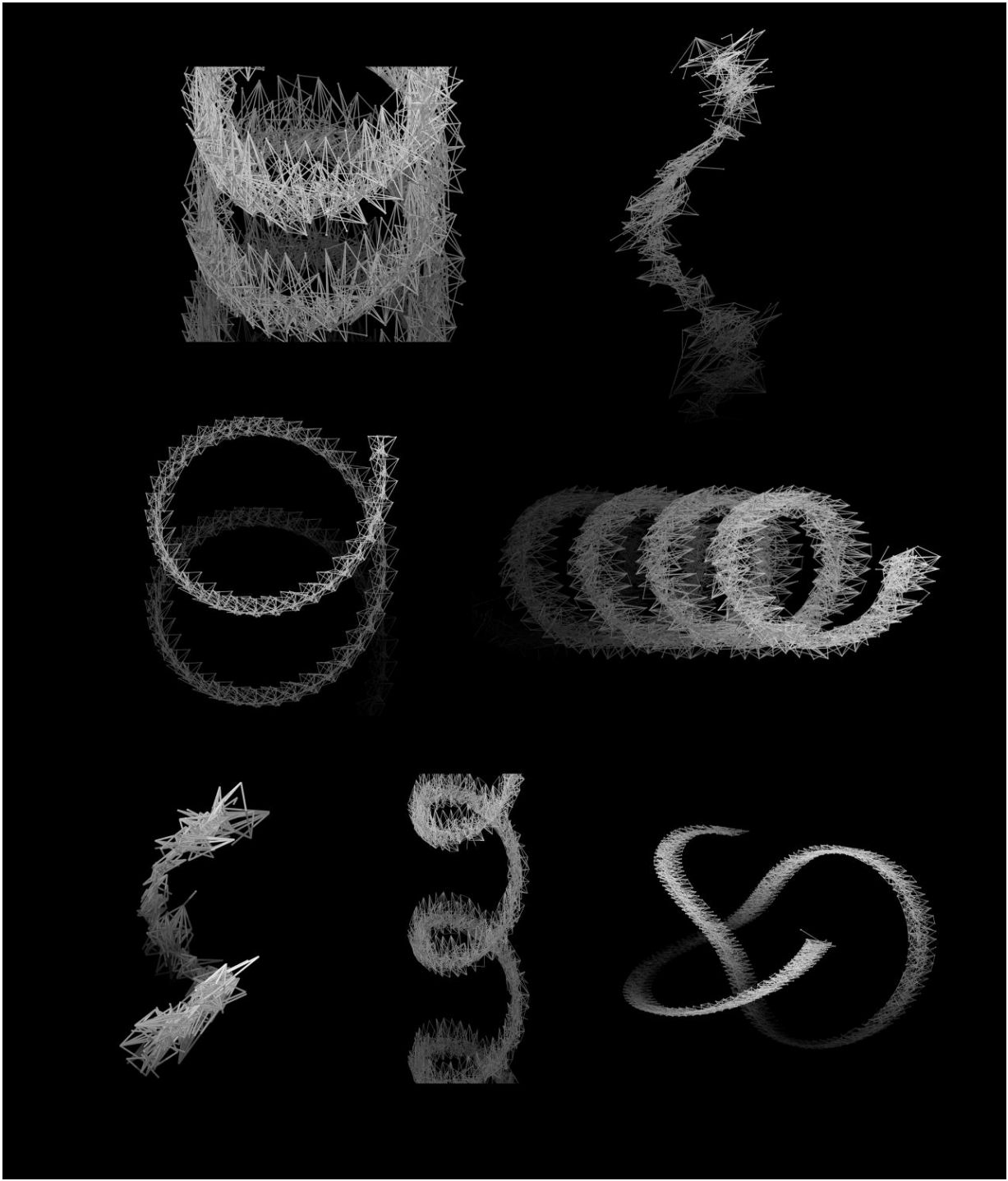

**Figure S5.** Gallery of various designs, showing the architectural complexity and variety that can be achieved using the algorithm. Different placement methods are used to shift/transform the positioning of graphs in space at each generative step.

**Supplementary Tables**

**Table S1.** Summary of deep graph generators in the literature, with a brief description of each.

| Type | Model | Training Feature | Main Characteristic |
|---|---|---|---|
| **Variational autoencoder (VAEs)** | GraphVAE (2018)[6] | Node/Edge | One of the most basic VAE graph generator |
| | Graphite (2018)[7] | Node | Applies an iterative graph refinement strategy |
| | CGVAE (2018)[8] | Node/Edge | Adapts graph linearization for permutation invariance and generation quality |
| | JT-VAE (2018)[9] | Node/Edge | Utilizes hierarchical chemical substructures as building blocks for molecule generation |
| | DGVAE (2020)[10] | Node | Incorporates balanced-graph-cut |
| **Generative adversarial network (GANs)** | NetGAN (2018)[11] | - | Combines random walks for graph generation |
| | CondGen (2019)[12] | - | Enables flexible semantic graph conditioning |
| | LGGAN (2020)[13] | Node | Generates labeled graphs |
| | TSGG-GAN (2020)[14] | Node | Considers time series for graph data |
| **Autoregressive models (ARs)** | GraphRNN (2018)[15] | - | Applies both node and edge sequence updates, and random breadth-first search for permutation |
| | GRAN (2018)[16] | - | Generates blocks of graphs each time step to improve generation quality and efficiency |
| | GraphGEN (2020)[17] | Node/Edge | Allows domain-agnostic data input, incorporate node and edge features for training, provides comprehensive evaluation metrics |
| | BiGG (2020)[18] | - | Integrates binary tree for embedding to save memory consumption |
| **Diffusion Models** | EDP-GNN (2020)[19] | Node/Edge | Uses score matching for permutation invariance |
| | GraphGDP (2022)[20] | Node/Edge | Uses position-enhanced score matching network to better capture graph topology information |
| | NVDiff (2022)[21] | Node/Edge | Combines VAE with score matching to reduce input dimension for computational efficiency |
| | GSDM (2022)[22] | Node | Applies spectrum-based model instead of full adjacency matrix |
| | DiGress (2022)[23] | Node/Edge | Enables conditioning for structural and spectral graph features |
| More for molecule generation | ORGAN (2017)[24], MolGAN (2018)[25], MolMP and MolRNN (2018)[26], SD-VAE (2018)[27], GraphNVP (2019)[28], GraphAF (2020)[29], MoFlow (2020)[30], EDM (2022)[31] | | |

**Table S2**: Parameters used in the sparse analog diffusion model

| Parameter | Value |
|---|---|
| U-net channels | 128 |
| Multipliers | [1, 2, 4] |
| Factors | [4,4] |
| Number of ResNet blocks | [2, 2] |
| Attention depths | [1, 1] |
| Attention heads | 8 |
| Attention multiplier | 2.0 |
| Conditioning embedding dimension $d_C$ | 256 |

**Table S3**: Parameters used in the diffusion model with full adjacency matrix

| Parameter | Value |
|---|---|
| U-net channels | 256 |
| Multipliers | [1, 2, 4] |
| Factors | [4, 4] |
| Number of ResNet blocks | [3, 3] |
| Attention depths | [1,1] |
| Attention heads | 8 |
| Attention multiplier | 2.0 |
| Conditioning embedding dimension $d_C$ | 256 |

**Table S4**: Parameters used in the autoregressive multi-headed transformer model with full adjacency matrix

| Parameter | Value |
|---|---|
| Decoder dimension | 512 |
| Depth | 12 |
| Head dimension/number of heads | 64/16 |
| Feed forward multiplier | 4.0 |
| Conditioning embedding dimension $d_C$ | 64 |

**Supplementary references**

1. Hamilton, W. L., Brachman, R. J., Rossi, F. & Stone, P. Graph Representation Learning.

2. Kipf, T. N. & Welling, M. *Variational Graph Auto-Encoders*. (2016).

3. Goodfellow, I. J. *et al.* Generative Adversarial Networks. (2014).

4. Wu, L., Cui Jian Pei, P. & Zhao Eds, L. *Graph Neural Networks: Foundations, Frontiers, and Applications*.

5. Karras, T., Aittala, M., Aila, T. & Laine, S. Elucidating the Design Space of Diffusion-Based Generative Models. (2022).

6. Simonovsky, M. & Komodakis, N. GraphVAE: Towards Generation of Small Graphs Using Variational Autoencoders. (2018).

7. Grover, A., Zweig, A. & Ermon, S. Graphite: Iterative Generative Modeling of Graphs. (2018).

8. Liu, Q., Allamanis, M., Brockschmidt, M. & Gaunt, A. L. Constrained Graph Variational Autoencoders for Molecule Design. (2018).

9. Jin, W., Barzilay, R. & Jaakkola, T. Junction Tree Variational Autoencoder for Molecular Graph Generation. (2018).

10. Li, J. *et al.* Dirichlet Graph Variational Autoencoder. (2020).

11. Bojchevski, A., Shchur, O., Zügner, D. & Günnemann, S. NetGAN: Generating Graphs via Random Walks. (2018).

12. Yang, C., Zhuang, P., Shi, W., Luu, A. & Li, P. *Conditional Structure Generation through Graph Variational Generative Adversarial Nets*.

13. Fan, S. & Huang, B. Labeled Graph Generative Adversarial Networks. (2019).

14. Yang, S., Liu, J., Wu, K. & Li, M. Learn to Generate Time Series Conditioned Graphs with Generative Adversarial Nets. (2020).

15. You, J., Ying, R., Ren, X., Hamilton, W. L. & Leskovec, J. GraphRNN: Generating Realistic Graphs with Deep Auto-regressive Models. (2018).

16. Liao, R. *et al.* Efficient Graph Generation with Graph Recurrent Attention Networks. (2019).

17. Goyal, N., Jain, H. V. & Ranu, S. GraphGen: A Scalable Approach to Domain-agnostic Labeled Graph Generation. in *The Web Conference 2020 - Proceedings of the World Wide Web Conference, WWW 2020* 1253–1263 (Association for Computing Machinery, Inc, 2020). doi:10.1145/3366423.3380201.

18. Dai, H., Nazi, A., Li, Y., Dai, B. & Schuurmans, D. Scalable Deep Generative Modeling for Sparse Graphs. (2020).

19. Niu, C. *et al.* Permutation Invariant Graph Generation via Score-Based Generative Modeling. (2020).

20. Huang, H., Sun, L., Du, B., Fu, Y. & Lv, W. GraphGDP: Generative Diffusion Processes for Permutation Invariant Graph Generation. (2022).

21. Chen, X., Li, Y., Zhang, A. & Liu, L. NVDiff: Graph Generation through the Diffusion of Node Vectors. (2022).

22. Luo, T., Mo, Z. & Pan, S. J. Fast Graph Generation via Spectral Diffusion. (2022).
11